# Standards for Language Resources

Nancy IDE
Department of Computer Science
Vassar College
Poughkeepsie, New York 12604-0520 USA
ide@cs.vassar.edu

Laurent ROMARY
Equipe Langue et Dialogue
LORIA/INRIA
54506 Vandoeuvre-lès-Nancy
romary@loria.fr

## Abstract

The goal of this paper is two-fold: to present an abstract data model for linguistic annotations and its implementation using XML, RDF and related standards; and to outline the work of a newly formed committee of the International Standards Organization (ISO), ISO/TC 37/SC 4 Language Resource Management, which will use this work as its starting point. The primary motive for presenting the latter is to solicit the participation of members of the research community to contribute to the work of the committee.

## Introduction

The goal of this paper is two-fold: to present an abstract data model for linguistic annotations and its implementation using XML, RDF and related standards; and to outline the work of a newly formed committee of the International Standards Organization (ISO), ISO/TC 37/SC 4 Language Resource Management, which will use this work as its starting point. The primary motive for presenting the latter is to solicit the participation of members of the research community to contribute to the work of the committee.

The objective of ISO/TC 37/SC 4 is to prepare international standards and guidelines for effective language resource management in applications in the multilingual information society. To this end, the committee will develop principles and methods for creating, coding, processing and managing language resources, such as written corpora, lexical corpora, speech corpora, dictionary compiling and classification schemes. The focus of the work is on data modeling, markup, data exchange and the evaluation of language resources other than terminologies (which have already been treated in ISO/TC 37). The worldwide use of ISO/TC 37/SC 4 standards should improve information management within industrial, technical and scientific environments, and increase efficiency in computer-supported language communication.

The standardization of principles and methods for the collection, processing and presentation of language resources requires a distinct type of standardization activity. Basic standards should be produced with wide-ranging applications in view. In the area of language resources, for instance, these standards should provide various technical committees of ISO, IEC and other standardizing bodies with the groundwork for building more precise standards for language resource management. ISO/TC 37/SC 4 will liaison with ISLE (International Standards for Language Engineering), which has implemented various recent efforts to integrate EC and US efforts for language resources. Where possible, these and other standards set up in EAGLES will be incorporated into the ISO standards. ISO/TC 37/SC 4 will also broaden the work of EAGLES/ISLE by including languages (e.g. Asian languages) that are not currently covered by EAGLES/ISLE standards.

We are aware that standardization is a difficult business, and that many members of the targeted communities are skeptical about imposing any sort of standards at all. There are two major arguments against the idea of standardization for language resources. First, the diversity of theoretical approaches to, in particular, the annotation of various linguistic phenomena suggests that standardization is at least impractical, if not impossible. Second, it is feared that vast amounts of existing data and processing software, which may have taken years of effort and considerable funding to develop, will be rendered obsolete by the acceptance of new standards by the community. To answer both of these concerns, we stress that the efforts of the committee are geared toward defining abstract models and general frameworks for creation and representation of language resources that should, in principle, be abstract enough to accommodate diverse theoretical approaches. The model so far developed in ISO TC/37 for terminology, which has informed and been informed by work on representation schemes for dictionaries and other lexical data (Ide, *et al.,* 2000) and syntactic annotation (Ide and Romary, 2001) demonstrates that this is not an unrealizable goal. Also, by situating all of the standards development squarely in the framework of XML and related standards such as RDF, we hope to ensure not only that the standards developed by the committee provide for compatibility with established and widely accepted web-based technologies, but also that transduction from legacy formats into XML formats conformant to the new standards is feasible.

At present, we feel that language professionals and standardization experts are not sufficiently aware of the standardization efforts being undertaken by ISO/TC 37/SC 4. Promoting awareness of future activities and rising problems, therefore, will be a crucial factor in the future success of the committee, and will be required to ensure widespread adoption of the standards it develops. An even more critical factor for the success of the committee's work is to involve, from the outset, as many and as broad a range of potential users of the standards as possible. This paper serves in part as a call for participation to the linguistics and computational linguistics research communities.

In the following sections, we describe the principles and architecture of a general framework for annotations that can serve to define an abstract format capable of representing the range of annotation types. We then provide examples in which the format is applied to specific annotations types.

# 1 Representing linguistic annotation

The goal of our work is to define a model for linguistic annotation that can (a) be instantiated in a standard representational format; and (b) can serve as a pivot format into and out of which proprietary formats can be transduced, in order to enable comparison and merging, as well as operation on the data by common tools, as shown in Figure 1.

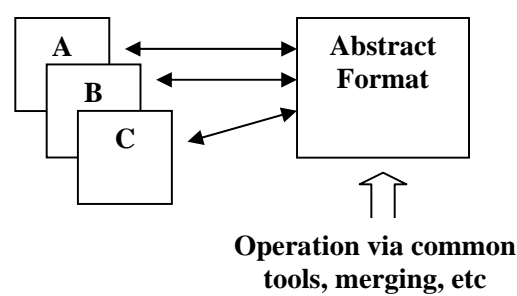

Figure 1. Overall architecture for handling multiple formats ("A", "B", and "C")

To accomplish this, it is necessary to identify a consistent underlying *data model* for data and its annotations. A data model is a formalized description of the data objects (in terms of composition, attributes, class membership, applicable procedures, etc.) and relations among them, independent of their instantiation in any particular form. A data model capable of capturing the structure and relations in diverse types of data and annotations is a pre-requisite for developing a common corpus-handling environment: it impacts the design of annotation schema, encoding formats and data architectures, and tool architectures.

## 1.1 Abstract model for annotation

At its highest level of abstraction, an annotation is a set of data or information (in our case, linguistic information) that is associated with some other data. The latter is what could be called "primary" data (e.g., a part of a text or speech signal, etc.), but this need not be the case; consider, for example, the alignment of parallel translations, where the "annotation" is a link between two primary data objects (the aligned texts). Typically, primary data objects are represented by "locations" in an electronic file, for example, the span of characters comprising a sentence or word, or a point at which a given temporal event begins or ends (as in speech annotation). As such, at the base primary data objects are relatively simple in their structure; more complex data objects may consist of a list or set of contiguous or non-contiguous locations. Annotation objects, on the other hand, often have a more complex internal structure: syntactic annotation, for example, may be expressed as a tree structure, and may include more elemental annotations such as dependency relations (which is itself an annotation relating two objects, where the relation is directional (dependent-to-head)).

Thus, we can conceive of, an annotation as a one- or two-way link between an annotation object and a point (or a list/set of points) or span (or a list/set of spans) within a base data set. Links may or may not have a semantics--i.e., a type--associated with them. Points and spans in the base data may themselves be objects, or sets or lists of objects. This abstract formulation can serve as the basis for defining a general model for linguistic annotation that can be realized in some representational format. However, we first make several observations:

- the model assumes a fundamental linearity of objects in the base,[1] e.g., as a time line (speech); a sequence of characters, words, sentences, etc.; or pixel data representing images;
- the *granularity* of the data representation and encoding is critical: it must be possible to uniquely point to the smallest possible component (e.g., character, phonetic component, pitch signal, morpheme, word, etc.);

---

[1] Note that this observation applies to the *fundamental* structure of stored data. Because the targets of a relation may be either individual objects, or sets or lists of objects, information with more than one dimension is accommodated.

- an annotation scheme must be mappable to the structures defined for annotation objects in the model;
- an encoding scheme must be able to capture the object structure and relations expressed in the model, including class membership and inheritance, therefore requiring a sophisticated means to specify linkage within and between documents;
- it is necessary to consider the logistics of identifying spans by enclosing them in start and end tags (thus enabling hierarchical grouping of objects in the data itself), vs. explicit addressing of start and end points;
- it must be possible to represent objects and relations in some (fairly straightforward) form that is both usable by a variety of tools and prevents information loss;
- ideally, it should be possible to represent the objects and relations in a variety of formats suitable to different tools and applications.

## 1.2   A closer look at annotations

As noted above, annotation objects may be relatively complex. In order to define a generic model for linguistic annotation, it is necessary to consider the representational needs for annotations themselves.

Linguistic annotation can be represented as a graph of elementary *structural nodes* to which one or more *information units* are attached. The distinction between the structure of annotations and the informational units of which it is comprised is, we feel, critical to the design of a truly general model for annotations. Annotations may be structured in several ways; perhaps the most common structure is hierarchical. For example, phrase structure analyses of syntax are structured as trees; in addition, hierarchy is often used to break annotation information into sub-components, as in the case of lexical and terminological information,

There are several special relations *among* annotations that must be represented in the model, including the following:

1   Parallelism: two or more annotations refer to the same data object;

2   Alternatives: two or more annotations comprise a set of mutually exclusive alternatives (e.g., two possible part-of-speech assignments, before disambiguation);

3   Aggregation: two or more annotations comprise a list or set that should be taken as a unit.

Information units or *data categories* provide the semantics of the annotation. Data categories are the most theory and application-specific part of an annotation scheme. We do not attempt to define the relevant data categories for given types of annotation. Rather, we propose the development of a Data Category Registry to provide a framework in which the research community can formally define data categories for reference and use in annotation, Data categories would be defined using RDF schemas to formalize the properties and relations associated with each. Note that RDF descriptions function much like class definitions in an object-oriented programming language: they provide, effectively, templates that describe how objects may be instantiated, but do not constitute the objects themselves. Thus, in a document containing an actual annotation, several objects with the same type may be instantiated, each with a different value. The RDF schema ensures that each instantiation is recognized as a sub-class of more general classes and inherits the appropriate properties.

A formally defined set of categories will have several functions: (1) it will provide a precise semantics for annotation categories that can be either used "off the shelf" by annotators; (2) it will provide a set of reference categories onto which scheme-specific names can be mapped; and (3) it will provide a point of departure for definition of variant or more precise categories. Thus the overall goal of the Data Category Registry is not to impose a specific set of categories, but rather to ensure that the semantics of data categories included in annotations (whether they exist in the Registry or not) are well-defined and understood.

## 0.2   The Generic Mapping Tool

We instantiate the abstract format for annotations using the eXtensible Markup Language (XML). XML is the emerging standard for data representation and exchange on the World Wide Web (Bray, *et al.,* 1998). Although at its most basic level XML is a document markup language directly derived from SGML (i.e., allowing tagged text (elements), element nesting, and element references), various features and extensions of XML (including XSLT, RDF, etc.) make it a powerful tool for data representation and access.

The model is instantiated by the Generic Mapping Tool (GMT), which includes the following tags:

- `<struct>` represents a structural node in the annotation. `<struct>` elements may be recursively nested at any level to reflect a tree structure for the annotation. Attributes include
  - *type* : annotation type (e.g., "syntax"), where necessary or desirable;
  - *ID*: unique identifier for the node
  - *ref* : node this `<struct>` node represents (for implicit structures)
- `<feat>` (feature) is used to provide information attached to the node represented by the enclosing `<struct>` element. A *type* attribute on the `<feat>` element identifies the data category of the feature. The tag may contain a string that provides an appropriate value for the data category (e.g., for *type=CAT* the value might be "NP") or `<feat>` can be recursively refined to describe complex structures. Alternatively, it may point via a *target* attribute to an object in another document that provides the value. Note that this allows the possibility for generating a single instantiation of an annotation value in a separate document that can be referenced as needed within the annotation document.
- `<alt>` is used to provide one or more alternative annotations, where necessary.
- `<rel>` is used to point to a non-contiguous related element, e.g., to identify dependencies explicitly by pointing to the

related `<struct>` node.

- `<seg>` points to the data to which the annotation applies. We assume the use of *stand-off annotation*[2]—i.e., annotation that is maintained in a document separate from the primary (annotated) data—as first defined in the Corpus Encoding Standard (CES) (Ide, 1998a, b) and subsequently adopted by the research community. A *target* attribute on the `<seg>` element uses XML Pointers (Xpointer) (Daniel, *et al.,* 2001) to specify the location of the relevant data.

- `<brack>` is used to group information to be regarded as a unit.

The GMT is sufficiently powerful to represent the information across all annotation types. We have already demonstrated its applicability to terminological and lexical information (Ide, *et al.,* 2000) and syntactic annotation (Ide and Romary, 2001); we provide additional examples below. Existing formats (whether or not in XML) can be mapped to the GMT, in order to enable merging, comparison, and manipulation via common tools. The GMT version can then be re-mapped to the original formats for use in in-house tools and applications. etc.

## 2    Examples

### 2.1    Morpho-syntactic annotation

We illustrate a simple application of the framework presented above for the domain of morpho-syntactic annotation. Morpho-syntactic annotation involves the identification of word classes over a continuous stream of word tokens. The annotations may refer to the segmentation of the input stream into word tokens, but may also involve grouping together sequences of tokens or identifying sub-token units (or morphemes), depending on the language under consideration and, in particular, the definitions of "word" and "morpheme" as applied to this language. The description of word classes may include one or several features such as syntactic category, lemma, gender, number etc., which is again dependent on the language being analyzed.

Morpho-syntactic annotation can be represented by a single type of structural node (named W-level) representing a word-level structure unit organized hierarchically. One or several information units are associated with each structural node.

For the purposes of illustration, we identify the following data categories (in practice these would be defined in reference to the Data Category Registry):

- /lemma/: contains or points to a reference word form for the token or sequence of tokens being described;
- /part of speech/: a reference to a morpho-syntactic category;
- /confidence/: a confidence level assigned by the manual or automatic annotator in ambiguous cases.
- /gender/: the grammatical gender information associated with a word token or a sequence of word tokens;
- /number/: the grammatical gender information associated with a word token or a sequence of word tokens;
- /tense/: the grammatical tense information associated with a word token or a sequence of word tokens;
- /person/: the grammatical person information associated with a word token or a sequence of word tokens.

The following provides an example of the morpho-syntactic annotation of the sentence "Paul aime les croissants" in the GMT format:[3]

```
<struct type="MSAnnot">
  <struct type="W-level">
    <feat type="lemma">Paul</feat>
    <feat type="pos">PNOUN</feat>
    <seg target="#w1"/>
  </struct>
  <struct type="W-level">
    <feat type="lemma">aimer</feat>
```

---

```
    <feat type="pos">VERB</feat>
    <feat type="tense">present</feat>
    <feat type="person">3</feat>
    <seg target="#w2"/>
 </struct>
 <struct type="W-level">
    <feat type="lemma">le</feat>
    <feat type="pos">DET</feat>
    <feat type="number">plural</feat>
    <seg target="#w3"/>
 </struct>
 <struct type="W-level">
    <feat type="lemma">croissant</feat>
    <feat type="pos">NOUN</feat>
    <feat type="number">plural</feat>
    <seg target="#w4"/>
 </struct>
</struct>
```

Note that there is no limit to the number of information units that may be associated with a given structural node (as opposed to the text based representations that are usually provided by available POS taggers). It is also possible to structure the annotations by embedding `<feat>` elements to reflect a more complex feature-based annotation, or by pointing to a lexical entry providing the information,

In some cases, the morpho-syntactic annotation of a word or sequence of words requires a hierarchy of word level structures (e.g., when a word token results from the combination of several morphemes that must be annotated independently). For example, some occurrences of the token "du" in French can be analyzed as the fusion of the preposition "de" with the determiner "le" (as in "la queue *du* chat"). This is handled by embedding word-level structures as follows:

```
<struct type="W-level">
  <seg target="#w1"/>
  <struct type="W-level">
      <feat type="lemma">de</feat>
      <feat type="pos">PREP</feat>
  </struct>
  <struct type="W-level">
      <feat type="lemma">le</feat>
      <feat type="pos">DET</feat>
  </struct>
</struct>
```

Conversely, annotation of compound words may involve associating a simple lemma to a sequence of word tokens at the surface level. In this case, the lemma is attached to the higher level of embedding and reference to the source is given at the leaves of the hierarchy, as in the following representation of the compound "pomme de terre" in French :

```
<struct type="W-level">
  <feat type="lemma">
          pomme_de_terre</feat>
  <feat type="pos">NOUN</feat>
  <struct type="W-level">
      <seg target="#w1"/>
      <feat type="lemma">pomme</feat>
      <feat type="pos">NOUN</feat>
  </struct>
  <struct type="W-level">
      <seg target="#w2"/>
      <feat type="lemma">de</feat>
      <feat type="pos">PREP</feat>
  </struct>
  <struct type="W-level">
      <seg target="#w3"/>
      <feat type="lemma">terre</feat>
      <feat type="pos">NOUN</feat>
  </struct>
</struct>
```

The ability to specify a hierarchical structure where needed enables specification of the level of granularity required. This is especially critical for a representation scheme, since the granularity of the segmentation in (or associated with) the primary data may not directly correspond to the level of granularity required for the annotation.

### 2.1.1 Alternatives

Morpho-syntactic annotation can be used to illustrate the representation of both structural and informational alternatives, which arises when a given word token is associated with two or more word classes. For example, the French word "bouche" which can be derived both from the verb "boucher" and the noun "bouche", which can be represented as follows:

```
<struct type="W-level">
  <seg target="#w1"/>
  <alt>
    <feat type="lemma">boucher</feat>
    <feat type="pos">VERB</feat>
    <feat type="tense">present</feat>
    <feat type="confidence">0.4</feat>
  </alt>
  <alt>
    <feat type="lemma">bouche</feat>
    <feat type="pos">NOUN</feat>
    <feat type="confidence">0.6</feat>
  </alt>
```

```
    </struct>
```

## 2.2   Relating annotation levels

As noted above, we assume the use of stand-off annotation; that is, an annotated corpus is represented as a lattice of stand-off annotation documents pointing to a primary source or intermediate annotation levels. However, depending on the point of view, the relations between various annotation levels can be more or less explicit. It is possible to identify three major ways to relate different levels of annotation: temporal anchoring, event-based anchoring, and object-based anchoring.

Temporal anchoring associates positional information to each structural level. This positional information is typically represented as a pair of numbers expressing the starting point and ending point of the segment being described. To do so in our framework, we introduce two attributes for the `<seg>` element:

- /startPosition/: the temporal or offset position of the beginning of the current structural node;
- /endPosition/: the temporal or offset position of the end of the current structural node.

For example, the following associates a phonetic transcription with a given portion of a primary text:

```
    <struct type="phonetic">
      <seg startsAt="2300"
          endsAt="3200"/>
      <feat type="phone">iy</feat>
    </struct>
```

We also define an event-based anchoring, which effectively introduces a structural node to represent a location in the text, to which all annotations for the object at that location can refer. This strategy is useful in two cases:

- Situations where it is not possible or desirable to modify the primary data by inserting markup to identify specific objects or points in the data (e.g., speech annotation, associated with a speech signal, or in general any "read-only" data).
- Primary data marked with "milestones", such as time stamps in speech data, where spans across the various milestones must be identified. In this case, the `<struct>` elements represent the markup for segmentation (e.g., segmentation into words, sentences, etc.).

To represent this, we introduce a specific type of structural node, named *landmark*, which is referred to by annotations for the defined span, as follows:

```
    <struct type="landmark">
        <seg startsAt="2300"
            endsAt="3200"/>
    </struct>
```

The annotation graph (AG) formalism (Bird and Liberman, 2001) was explicitly designed to deal with time-stamped data. We can represent annotation graphs in the GMT as shown in Figure 2. We feel that the GMT representation is more general, for the following reasons:

- The AG formalism reifies the "arc" and treats it as a special

```
<annotation>
<arc><source  id="0"   offset="0"/><label  att_1="P"  att_2="h#"/><target  id="1"
offset="2360"/></arc>
<arc><source  id="1"   offset="2360"/><label  att_1="P"  att_2="sh"/><target  id="2"
offset="3270"/></arc>
<arc><source  id="2"   offset="3270"/><label  att_1="P"  att_2="iy"/><target  id="3"
offset="5200"/></arc>
<arc><source  id="1"   offset="2360"/><label  att_1="W"  att_2="she"/><target  id="3"
offset="5200"/></arc>
<arc><source  id="3"   offset="5200"/><label  att_1="P"  att_2="hv"/><target  id="4"
offset="6160"/></arc>
</annotation>
```

```
<arc><source  id="4"  offset="6160"/><label  att_1="P"  att_2="ae"/><target  id="5"
offset="8720"/></arc>
<arc><source  id="5"  offset="8720"/><label  att_1="P"  att_2="dcl"/><target  id="6"
offset="9680"/></arc>
<arc><source  id="3"  offset="5200"/><label  att_1="W"  att_2="had"/><target  id="6"
offset="9680"/></arc>
<arc><source  id="6"  offset="9680"/><label  att_1="P"  att_2="y"/><target  id="7"
offset="10173"/></arc>
<arc><source  id="7"  offset="10173"/><label  att_1="P"  att_2="axr"/><target  id="8"
offset="11077"/></arc>
<arc><source  id="6"  offset="9680"/><label  att_1="W"  att_2="your"/><target  id="8"
offset="11077"/></arc>
</annotation>
```

Figure 2a. XML instantiation of an annotation graph

```
<struct type="landmarkDesc>                      <struct type="phone">
    <struct type="landmark" id="0">                 <startsAt target="#1"/>
        <position>0</position>                      <endsAt target="#2"/>
    </struct>                                        <phone>sh</phone>
    <struct type="landmark" id="1">             </struct>
        <position>2360</position>               <struct type="phone">
    </struct>                                        <startsAt target="#2"/>
    <struct type="landmark" id="2">                 <endsAt target="#3"/>
        <position>5200</position>                   <phone>iy</phone>
    </struct>                                    </struct>
    ...                                         ...
</struct>                                    </struct>

<struct type="phoneticAnnot">                   <struct type="morphAnnot">
    <struct type="phone">                           <struct type="w">
        <startsAt target="#0"/>                         <startsAt target="#0"/>
        <endsAt target="#1"/>                           <endsAt target="#3"/>
        <phone>h#</phone>                               <source>she</source>
    </struct>                                        </struct>
                                                ...
                                            </struct>
```

Figure 2b. Annotation graph representation in GMT

The third mechanism, object-based anchoring, enables pointing from a given level to one or several structural nodes at another level. This mechanism is particularly useful to make dependencies between two or more annotation levels explicit. For example, syntactic annotation can refer directly to the relevant nodes in a morpho-syntactically annotated corpus, in order, for example, to identify the correct NP "le chat" in "la queue du chat", as shown below:

```
<!-- Morphosyntactic level -->
<struct type="W-level">
    <seg target="#w3">
    <struct type="W-level">
        <seg target="#w3.1">
        <feat type="lemma">de</feat>
        <feat type="pos">PREP</feat>
    </struct>
        <struct type="W-level">
```

```
        <seg target="#w3.2">
        <feat type="lemma">le</feat>
        <feat type="pos">DET</feat>
        <feat type="gender">masc</feat>
    </struct>
    </struct>
    <struct type="W-level">
        <seg target="#w4">
        <feat type="lemma">chat</feat>
        <feat type="pos">NOUN</feat>
    </struct>
</struct>
<!-- Syntactic level (simplified) -->
<struct>
    <feat type="synCat">NP</feat>
    <seg targets="w3.2 w4"/>
</struct>
```

## Conclusion

The framework presented here for linguistic annotation is intended to allow for variation in annotation schemes while at the same time enabling comparison and evaluation, merging of different annotations, and development of common tools for creating and using annotated data. We have developed an abstract model for annotations that is capable of representing the necessary information while providing a common encoding format that can be used as a pivot for combining and comparing annotations, as well as an underlying format that can be manipulated and accessed with common tools. The details presented here provide a look "under the hood" in order to show the flexibility and representational power of the abstract scheme; however, the intention is that annotators and users of syntactic annotation schemes can continue to use their own or other formats with which they are comfortable, and translation into and out of the abstract format will be automatic.

Our framework for linguistic annotation is built around some relatively straightforward ideas: separation of information conveyed by means of structure and information conveyed directly by specification of content categories; development of an abstract format that puts a layer of abstraction between site-specific annotation schemes and standard specifications; and creation of a Data Category Registry to provide a reference set of annotation categories. The emergence of XML and related standards, together with RDF, provides the enabling technology. We are, therefore, at a point where the creation and use of annotated data and concerns about the way it is represented can be treated separately—that is, researchers can focus on the question of *what* to encode, independent of the question of *how* to encode it. The end result should be greater coherence, consistency, and ease of use and access for linguistic annotated data.